\documentclass[runningheads]{llncs}
\usepackage{graphicx}
\usepackage{comment}
\usepackage{amsmath,amssymb} %
\DeclareMathOperator*{\argmax}{argmax}
\usepackage{color}
\usepackage{wrapfig}

\usepackage{url}
\usepackage{hyperref}
\hypersetup{
    colorlinks=true,
    linkcolor=black,
    filecolor=magenta,      
    urlcolor=blue,
}
\usepackage[dvipsnames]{xcolor}
\usepackage{algorithm}
\usepackage{algorithmic}
\usepackage{array}

\begin{document}

\pagestyle{headings}
\mainmatter
\def\ECCVSubNumber{4871}  

\title{Inference Graphs for CNN Interpretation} 

\titlerunning{Inference Graphs for CNN Interpretation}
%
\author{Yael Konforti$^\dagger$\and
Alon Shpigler$^\dagger$ \and
Boaz Lerner\and Aharon Bar-Hillel}
\authorrunning{Konforti Y., Shpigler A., Lerner B., Bar-Hillel A.}
%
\institute{Ben-Gurion University of the Negev, Beer Sheva, Israel\\
\email{\{yaelkonf, alonshp\}@post.bgu.ac.il; 
\{boaz, barhille\}@bgu.ac.il}
\footnotetext{$^\dagger$ Equal contribution\\ Code for inference graphs algorithm released at \href{https://github.com/yaelkon/GMM-CNN}{github.com/yaelkon/GMM-CNN}}}
\maketitle

\begin{abstract}
Convolutional neural networks (CNNs) have achieved superior accuracy in many visual related tasks. However, the inference process
through intermediate layers is opaque, making it difficult to interpret
such networks or develop trust in their operation. We propose to model the network hidden layers activity using probabilistic models. The activity patterns in layers of interest are modeled as Gaussian mixture models, and transition probabilities between clusters in consecutive modeled layers are estimated.
Based on maximum-likelihood considerations, nodes and paths relevant for network prediction are chosen, connected, and visualized as an inference graph. 
We show that such graphs are useful for understanding the general inference process of a class, as well as  explaining decisions the network makes regarding specific images. 

\end{abstract}

\section{Introduction}
\label{sec:introduction}

Thanks to their impressive performance, convolutional neural networks (CNNs) are the leading architecture for tasks in computer vision~\cite{krizhevsky2012imagenet,he2016deep,huang2017densely}. 
However, due to their complex end-to-end training and architecture, understanding of their decision-making process and task assignment across hidden layers is lacking. 
This turns network interpretability into a difficult problem, and undermines usage of deep networks when high reliability and inference transparency are required. 
Understanding CNN reasoning by decomposing it into layer-wise stages can provide insights about cases of failure, and reveal weak spots in the network architecture, training scheme, or data-collection mechanism. In turn, these insights can lead to more robust networks and develop more trust in CNN decisions.

\begin{figure}[!t]
\centering

\includegraphics[width=0.85\columnwidth]{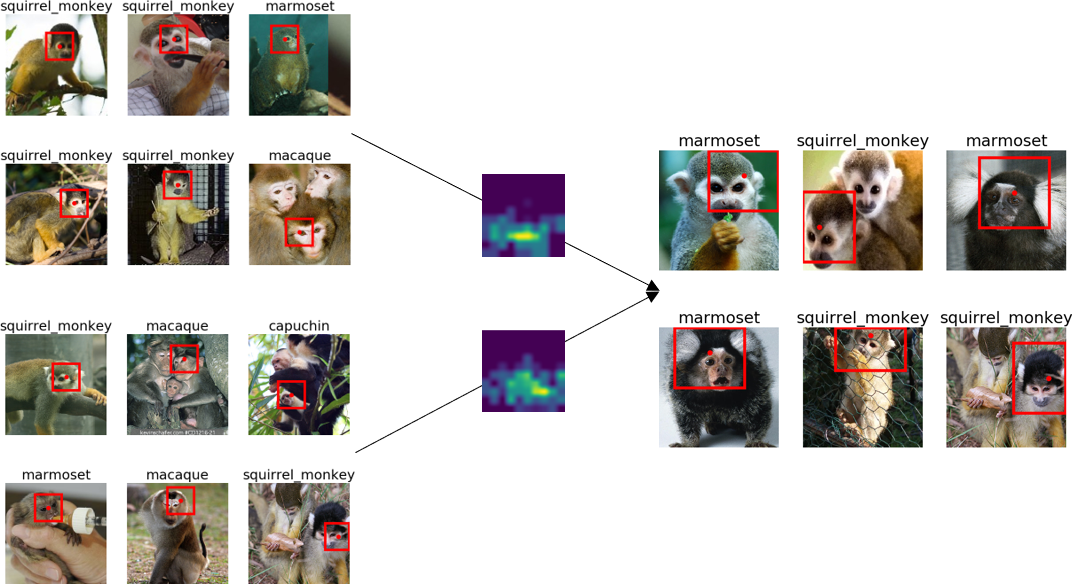}
 \caption{\footnotesize{\textbf{Visual words connected to form an inference path.} Two visual words in a lower network layer (left) (each is represented by six most typical images), representing the monkey eye (top) and face (bottom), explain a visual word in an upper network layer (right) that represents the monkey face. The heatmaps show spatial densities of lower-layer visual words in the receptive field of the higher-level word (see Section~\ref{subsec:cnn_res}).}} 
 \label{fig:intro}
\end{figure}

As we see it, enabling human understanding of the deep network inference process requires facing a main challenge of
transforming CNN activities into discrete representations amendable to human reasoning. 
Deep networks operate through a series of distributed layer representations.
Human language, however, is made up of discrete symbols, i.e., words, having meaning grounded by their reference in the world of objects, predicates, and their interrelations. 
The question of interest in this respect is: \textit{Can we convert distributed representations into a human-oriented language?} More technically, can we learn a dictionary of visual words and model their interrelationships, leading to an interpretable inference graph?



To address this, we suggest to describe the inference process of a network with probabilistic models. We model activity vectors in each layer as arising from a multivariate Gaussian mixture model (GMM). Layer activity in fully connected (FC) layers, or spatial location activity in convolutional layers, is associated with one of \textit{K} clusters (GMM components), each representing a visual word. Connections between visual words of consecutive layers are modeled using conditional probabilities. For a multilayer perceptron (MLP) network, a full model with efficient inference can be obtained using a hidden Markov model (HMM). For the convolutional layers, each spatial location has its own hidden variable and dependencies among visual words in consecutive layers are described using conditional probability tables.
Given a selected subset of images to be explained (either a specific image or images of an entire class), we describe the decision process of the network using an inference graph, representing the visual words used in different hidden layers and their probabilistic connections. As the full graph may contain thousands of visual words, a useful explanation has to find informative subgraphs, containing the most explanatory words w.r.t the network decision, and the likely paths connecting them. We suggest an algorithm for finding such graphs based on maximum-likelihood considerations. 

Our contributions are: (\romannumeral 1) a new approach for network interpretation, providing a formalism for probabilistic reasoning about inference processes in deep networks,
and (\romannumeral 2) a graph-mining algorithm and visual tools enabling inference understanding. Our suggested inference graph provides a succinct summary of the inference process performed on a specific class of images or a single image, as inference progresses through the network layers. 
An example of visual nodes and the inferred connection between them is shown in Fig.~\ref{fig:intro}.


\section{Related Work}\label{sec:relatedwork}
\textbf{Network visualization.} 
Several techniques have been suggested to visualize network behavior.
In activation maximization~\cite{erhan2009visualizing}, the input that maximizes the score of a given hidden unit is visualized by carrying out regularized gradient ascent optimization in the image space, 
as was applied to output class neurons~\cite{simonyan2013deep} and  intermediate layer neurons~\cite{Yosinski2015}. Another technique~\cite{simonyan2013deep} visualizes the gradient strength in the original image space for a specific example, providing a ``saliency map'' showing class score sensitivity to image pixels.
The idea was extended to an entire class of interest in~\cite{zhou2016learning}.
In~\cite{zeiler2014visualizing}, the role of neurons in intermediate layers was visualized by an inverse de-convolution network.
\newline
\textbf{Simplifying network representation.} 
Similar to our work, several works looked for categorizing features through clustering~\cite{liao2016learning,chen2019looks}. Liao et al.~\cite{liao2016learning} added a regularization term encouraging the network representation to form three kinds of clusters governed by examples, spatial locations, and channels. 
A similar approach was introduced for learning class discriminative clusters of spatial columns~\cite{chen2019looks}.
However, these approaches influence the trained network and trade accuracy for explainability, whereas ours finds meaningful inference explanations without interfering with the network learning process.               
\newline
\textbf{Modeling relationships between consecutive layer representations.} In CNNVis~\cite{liu2016towards}, neurons in each layer are clustered to form groups having similar activity patterns. For the clustering, a neuron was described using a $C$-dimensional vector of its average activity on each class $1,..,C$. A graph between clusters of subsequent layers was then formed based on the average weight strengths between cluster neurons. 
Note that this method clusters neurons, while we cluster activity vectors (of neuronal columns) across examples.
Olah et al.~\cite{olah2018building} proposed a tool for visualizing the network path for a single image. They decomposed each layer's activations into groups using matrix factorization, and connected groups from consecutive layers into a graph structure similar to~\cite{liu2016towards}.

\section{Method}\label{sec:method}

Inference graphs for an MLP, for which a full graphical model can be suggested, are presented in Section~\ref{subsub:fc_nets}. The more general case of a CNN is discussed in Section~\ref{subsubsec:CNN}, and its related graph-mining algorithm in Section~\ref{subsec:node_selelction}.
Models can be trained on the full set of network layers or on a subset, indexed by $l\in\{1,\dots,L\}$.

\subsection{Inference Graphs for MLPs}
\label{subsub:fc_nets}

\begin{figure}[t]
\centering
\includegraphics[width=0.83\columnwidth]{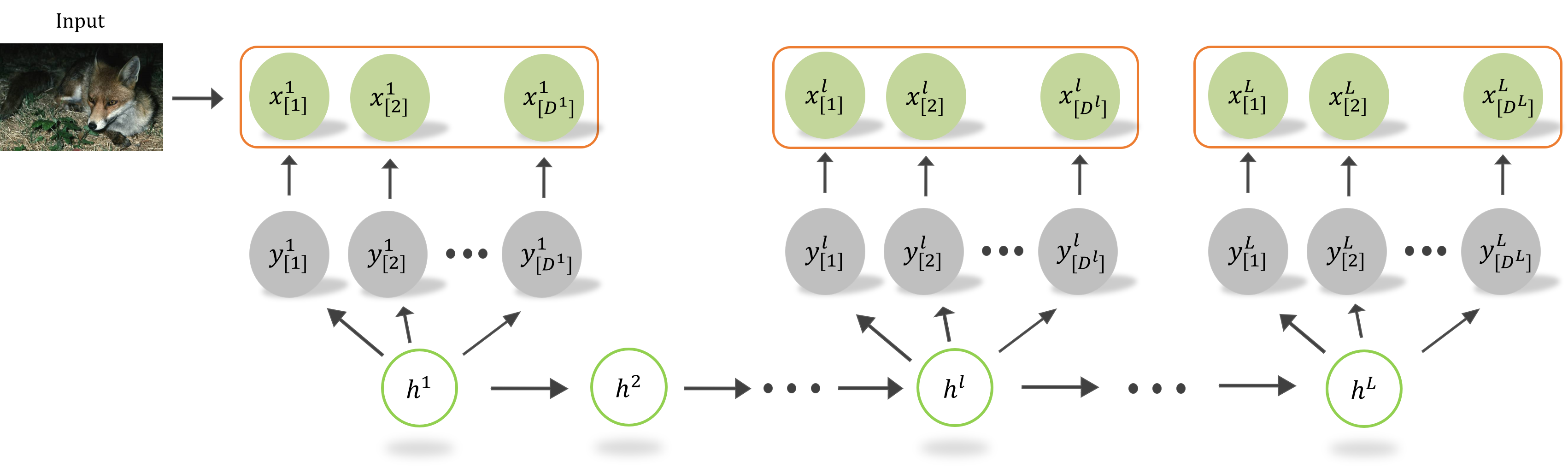}

\caption{ \textbf{HMM for MLP networks.} 
Orange rectangles represent post-ReLU layer activity. Activation $x^l[d]$ of neuron $d$ in layer $l$ is generated from a rectified Gaussian density, where
$x^l[d]$'s parent, $y^l[d]$, is a Gaussian density before rectification, and
$h^l$ is a hidden variable generating the hidden vector of multivariate Gaussians. 
}

\label{fig:HMM}
\end{figure}

The hidden layer activity of an MLP network composed of FC layers can be modeled by a single probabilistic graphical model with an HMM structure, enabling closed-form inference.
The model structure is shown in Fig.~\ref{fig:HMM}. 

The activation vector 
of the $l^{th}$ FC layer with $D^l$ neurons, $x^l =(x^l[1],..,x^l[D^l])$ $\in \mathbb{R}^{D^l}$, results from a mixture of $K^l$ components with a discrete hidden variable $h^l\in \{1,...,K^l\}$ denoting the component index a sample is assigned to, i.e. its cluster. To model the ReLU operation, each activation $x^l[d]$ is generated from a rectified Gaussian distribution~\cite{socci1998rectified}.
The conditional probability $P(x^l|h^l)$ is hence assumed to be a rectified multivariate Gaussian distribution with a diagonal covariance matrix. 
Connections between hidden variables in consecutive layers are modeled by a conditional probability table (CPT) $P(h^l|h^{l-1})$.

Using this generative model, an activity pattern for the network is sampled by three steps.
First, a path $(h^1,\dots,h^L)$ of hidden states is generated according to the transition probabilities 
$P(h^l=k|h^{l-1}=k')=t_{k,k'}^l$,
where $t^l\in \mathbb{R}^{K^l\times K^{l-1}}$ is a learned CPT. For notation simplicity, we define $h^0=\{\}$, so $P(h^1|h^0)$ is actually $P(h^1)$ parametrized by $P(h^1=k)=t_k^1$. 
After path generation, ''pre-ReLU'' Gaussian vectors $(y^1,\dots,y^L)$, with $y^l\in \mathbb{R}^{D^l}$, are generated based on the chosen hidden variables. A single variable $y^l[d]$ is formed according to

\begin{equation}
\label{eq:y|h}
P(y^l [d]|h^l=k) \sim \mathcal{N}(y^l [d]| \mu_{d,k}^l,\sigma_{d,k}^l),
\end{equation}

\noindent where $ \mu_{d,k}^l$ and $ \sigma_{d,k}^l$ are the mean and standard deviation of the $d$th element in the $k$th component of layer $l$.
Since the observed activity $x^l[d]$, generated as
$x^l[d]= max(y^l[d],0)$, is a deterministic function of $y^l[d]$, its conditional probability $P(x^l[d]|y^l[d]) $ can be written as 

\begin{equation}
\label{eq:xy}
	P(x^l [d]|y^l [d] ) = \bigg\{ 
	\begin{array}{c c}
	\delta_{x^l [d] = y^l [d]} &,y^l[d] > 0\\
	\delta_{x^l [d] = 0}\quad \qquad &,y^l[d] \leq 0

	\end{array} ,
\end{equation}

\noindent with $\delta_{(x=c)} $ as the Dirac delta function concentrating the distribution mass at $c$.

The full likelihood of the model is given by
\begin{equation}
P(X,Y,H|\Theta)=\prod_{l=1}^L P(h^l|h^{l-1}) P(y^l|h^l) P(x^l|y^l),
\label{eq:like}
\end{equation}
\noindent where $Y, H, X$ are tuples representing their respective variables across all layers (e.g., $H={\{h_l\}}_{l=1}^L$), and the equation's components are stated above.

\textbf{Training algorithm:} \label{subsubsec:fc_training_algo} In~\cite{lee2012algorithms}, the EM formulation was suggested for training a mixture of rectified Gaussians. We extended this idea to the HMM formulation in an online setting.
Following~\cite{cappe2009line}, the online EM algorithm tracks the sufficient statistics using running averages, and updates the model parameters using these statistics. 
Explicit update formulas in terms of the tracked sufficient statistics are presented in the Supplementary Material.


\subsection{Inference Graphs for CNNs}\label{subsubsec:CNN}
\textbf{Layer dictionaries}: In a CNN, the activation output of the $l$th convolutional layer is a tensor $X^l\in R^{H^l\times W^l\times D^l}$, where $H^l$, $W^l$, and $D^l$ correspond to the height, width, and number of maps, respectively.
We consider the activation tensor as consisting of $H^l \times W^l$  spatial column examples, $x_{p}^{l}\in R^{D^l}$, located at $p = (i,j)\in\{\{1,\dots,H^l\} \times \{1, \dots\, W^l\}\}$, and wish to model each such location as containing a separate visual word from a dictionary shared by all locations. 

The number of hidden variables (one per location) is much larger than that in an FC layer (where a single hidden variable per layer was used), and their connectivity pattern across layers is dense, leading to a graphical model with high induced width, but with infeasible exact inference. Hence, we turn to simpler model and training techniques that are 
scalable to the size and complexity of CNNs. 
In this model, spatial column $x_{p}^{l}$ is described as arising from a GMM of $K^l$ clusters, regarded as visual words forming the layer dictionary. Using a training image set $S_T =\{(I_{n}, y_n)\}_{n=1}^{N_{T}}$, the GMM is trained independently for each layer of interest. While each location $p$ in layer $l$ has a separate hidden random variable, $h_p^l$, the GMM parameters are shared across all the spatial locations of that layer.
After model training, the activity tensor of layer $l$ for an example $I$ can be mapped into a tensor $P\in R^{H^l\times W^l \times K^l}$, holding $P(h_{p}^l(I)=k)$. With slight abuse of notation, we say that $h^l_{p}(I)=k^*$ (an activation column of image $I$ in position $p$ is assigned to cluster $k^*$) iff $k^* = \argmax_k P(h^l_{p}(I)=k)$. 
Accordingly, visual word $k$ in layer $l$ is the cluster\break $C^l_k = \{(I,p), I \in S_T: h^l_p(I)=k\}$ containing activations over all positions for all images in the training set $S_T$, where cluster $k$ has the highest $P(h^l_{p}(I)=k)$.

When the CNN also contains global layers, these can be modeled using a GMM trained on the layer's activity vectors. This can be regarded as a degenerate case of convolutional layer modeling, where the number of spatial locations is one. Specifically, the output layer of the network, $X^L$, containing the $M$ class of predicted probabilities,
is modeled using a GMM of $M$ components. This GMM is not trained, and instead is fixed such that $\mu_{d,m} = 1$ for $d=m$ and $0$ otherwise, with a constant standard deviation of $\sigma_{d,m} = 0.1$. In this setting, cluster $m$ of the output layer contains 
images that the network predicts to be of class $m$. 

\textbf{Probabilistic connections between layer dictionaries:}
Transition probabilities between visual words in consecutive layers are modeled a-posteriori.
For two consecutive modeled layers $l'$ and $l$ ($l'<l$), the receptive field $R(p)$ of location $p$ in layer $l$ is defined as the set of locations  $\{q = p + o:  o\in O\}$ in layer $l'$ used in the computation of $x^l_p$. $O$ is a set of $\{(\Delta x, \Delta y)\}$ integer offsets. Using a validation sample 
$S_V=\{I_{n}\}_{n=1}^{N_{V}}$, we compute 
the co-occurrence matrix $N \in M^{K^{l}\times K^{l'}}$ between the visual words contained in the dictionaries of layers $l$ and $l'$, 
\begin{equation}
N(k,k') = \big|\{(I_{n},p,q): h^{l}_{p}(I_{n})=k, h^{l'}_{q}(I_{n})=k', q\in R(p)\}\big|.
\end{equation}

\noindent Using $N$, we can obtain the following first and second order statistics:
\begin{equation}
\label{eq:first_order}
    \hat{P}(h^{l} = k) = \frac{\sum_{j} N(k,j)}{\sum_{i,j} N(i,j)}
\end{equation}
\begin{eqnarray}
\label{eq:second_order}
 & \hat{P}\big(h^{l'}_q = k' | h^{l}_p = k, q\in R(p)\big) = \frac{N(k,k')}{\sum_{j} N(k,j)}  = \\ \nonumber
 & \frac{\big |\big\{(I_{n},p,q): h^{l}_{p}(I_{n}) = k, h^{l'}_{q}(I_{n})=k', q\in R(p)\big\}\big|}{|O| \cdot  \big|\big\{(I_{n},p): h^{l}_{p}(I_{n})=k\big\}\big|} = \frac{1}{|O|}\sum_{o \in O} \hat{P}(h^{l'}_{p+o} = k' | h^{l}_{p} = k).
\end{eqnarray}
The transition probabilities as defined above are abbreviated in the following discussion to $\hat{P}(h^{l'}  =  k' | h^{l} = k)$. These probabilities are averaged over specific positions in the receptive field, since modeling of position-specific transition probabilities separately would lead to proliferation in the parameters number. 

\textbf{Training algorithm}:
\label{subsubsec:cnn_training}
The GMM parameters $\Theta^l$ of layer $l$ are trained by associating a GMM layer to each modeled layer of the network. Since we do not wish to alter the network's behavior, the GMM gradients 
do not propagate towards lower layers of the network. 
 We considered two optimization approaches for training $\Theta^l$: 
\newline
(1) \textit{Generative loss} --- The optimization objective is to minimize the negative  log-likelihood function:
    \begin{equation}
    \label{eq:generative_loss}
       \mathcal{L}_{G}(X^l(I_n),\Theta^l) =
        -\sum_{p\in \{1,...,H^{l}\}\times \{1,...,W^{l}\}} \log \sum_{k=1}^{K^l} \pi^{l}_{k}G(x_p^l(I_n)|\mu^{l}_{k}, \Sigma^{l}_{k})
    \end{equation}

\noindent where $G$ is the Gaussian distribution function, and $\pi_k^l$ is the mixture probability of the $k$'th component in layer $l$.
\newline
(2) \textit{Discriminative loss} --- The probability tensor $P$ is summarized into a histogram of visual words $Hist^l(X^l(I_n))\in R^{K^l}$ using a global pooling operation. A linear classifier $\mathcal{W} \cdot Hist^l(X^l(I_n))$ is formed and optimized by minimizing a cross entropy loss, where $\mathcal{W}$ is the classifier weights vector,
\begin{equation}
 \label{eq:discriminative_loss}
\mathcal{L}_{D}(X^l(I_n),\Theta^l, y_n) =
 - \log{P(\hat{y_n}=y_n|\mathcal{W} \cdot Hist^l(X^l(I_n), \Theta^l))}
\end{equation}
and $\hat{y_n}$ is the predicted output after a softmax transformation.
Empirical comparison between these two approaches is given in Section~\ref{subsec:cnn_res}.

For ImageNet-scale networks,
full modeling of the entire network at once may require thousands of visual words per layer. Training such large dictionaries is not feasible with current GPU memory limitations (12GB for a TitanX). Our solution is to train a class-specific model, explaining network behavior for a specific class $m$ and its ``neighboring'' classes, i.e., all classes erroneously predicted by the network for images of class $m$. The set of neighboring classes is chosen based on the network's confusion matrix computed on the validation set. 
The model is trained on all training images of class $m$ and its neighbors.
\subsection{Graph Node Selection Algorithm}\label{subsec:node_selelction}

Consider a graph in which column activity clusters (i.e., visual words)\break ${\{C^l_k\}}_{l=1,k=1}^{L,K^l}$
are the nodes, and transition probabilities between clusters of consecutive layers quantify edges between the nodes. Typically, this graph contains thousands of nodes and, thus, is not feasible for human interpretation. However, specific subgraphs may have high explanatory value. Specifically, nodes (clusters) of the final layer $C^L_k$ in this graph represent images for which the network predicted a class $k$. To understand this decision, we evaluate clusters in the previous layer $C^{L-1}_{k'}$ using a score based on the transition probabilities $P(h^L=k|h^{L-1}=k')$. The step of finding such a set of ``explanatory" clusters in layer $L-1$ is repeated to lower layers. Below, we develop an iterative algorithm that using a validation subset of images $\Omega=\{I_n\}_{n=1}^N$ outputs a subgraph of the nodes that most ``explain'' the network decisions on $\Omega$, where ``explanation'' is defined in the maximum-likelihood sense. We first explain node selection for a single visual word in a single image, and then extend this notion to a full algorithm operating on multiple visual words and images.   
\subsubsection{Explaining a single visual word:}
\label{subec:SVW}
Consider an instance of a single visual word $h_p^l(I)=s$, derived from a column activity location $p$ in layer $l$ for image $I$. Given this visual word, we look for the visual words in $R(p)$ most contributing to its likelihood, given by (omitting the image notation $I$ in $h_p^l(I)$ for brevity):
\begin{gather}
\label{eq:SingleWord}
\scalebox{0.95}{$\begin{align}
&P\Big(h_p^l  = s \big|\big\{h_q^{l'}:q\in R(p)\big\}\Big) = \frac{P   \Big( \big\{h_q^{l'} :  q\in R(p)\big\} \big| h_p^l = s\Big) \cdot P(h_p^l = s)}{P\Big(\big\{h_q^{l'}:q\in R(p)\big\}\Big)} \quad \\ \nonumber
   &\approx \frac{\prod_{q \in R(p)} P(h_q^{l'
    } | h^l_p=s) \cdot P(h_p^l=s)}{\prod_{q \in R(p)} P(h_q^{l'
    })}.
    \end{align}$}
\end{gather}
In the last step, two simplifying assumptions were made: conditional independence over locations in the receptive field (nominator) and independence of locations (denominator).
Taking the logarithm, we decompose the expression:
\begin{gather}
\label{eq:A+C*log(lr)}
\scalebox{0.95}{$\begin{align}
   &\underbrace{\log P(h_p^l=s)}_\textrm{constant $A$} +\sum_{q \in R(p)}\log \frac{ P(h_q^{l'
    } | h^l_p=s) }{P(h_q^{l'})} = \\ \nonumber
       &A + \sum_{t=1,\dots,K^{l'}} \big|\big\{q:h_q^{l'}=t, q \in R(p)\big\}\big|\log \frac{P\big(h_q^{l'}=t|h_p^l=s, q \in R(p)\big)}{P(h_q^{l'}=t)}.
       \end{align}$}
\end{gather}

Denote by $Q_t^{l'}(I,p) = \big|\big\{q:h_q^{l'}=t, q \in R(p)\big\}\big|$ the number of times visual word $t$ appears in the receptive field of location $p$. 
We look for a subset of words $T\subset \{1,\dots,K^{l' }\}$, which contribute the most to the likelihood of $h_p^l=s$. Thus, the problem we solve is
\begin{equation}
\label{eq:optML}
\scalebox{0.95}{$
\max_{\underset{|T|=Z}{T}}\left\{ \sum_{t \in T} Q_t^{l'}(I,p) \log \frac{P\big(h_q^{l'}=t | h_p^l=s, q \in R(p)\big)}{P(h_q^{l'}=t)}\right\}, $}
\end{equation}
\noindent  The solution is obtained by choosing the first $Z$ words for which the  score 
\begin{equation}
\label{eq:S(I,s,t)}
\scalebox{0.95}{$
    S^{l'}(I,s,t)= Q_t^{l'}(I,p) \log \frac{P\big(h_q^{l'}=t|h_p^l=s, q \in R(p)\big)}{P(h_q^{l'}=t)}$}
\end{equation}
is the highest.
Intuitively, the score of the $t$th  visual word is the product of two terms, $Q_t^{l'}(I,p)$, which measures the word frequency in the receptive field, and $\log \frac{P\big(h_q^{l'}=t|h_p^l=s, q \in R(p)\big)}{P(h_q^{l'}=t)}$, which measures how likely it is to see word $t$ in the receptive field compared to seeing it in general. To compute the probabilities in the log term of the score, we use the estimations $\hat{P}(h^l=k)$ and $\hat{P}(h^{l'}=k'|h^l=k)$ made using Eqs.~\ref{eq:first_order} and \ref{eq:second_order}, respectively.

 \begin{algorithm}[t]
 \caption{\textbf{\label{alg:GraphBuild}}Inference graph building}
 {

 \textbf{\small{{Input:}}}{\small{
 CNN $CN$, a set $\Omega$ of images predicted by $CN$ to class $m$, network model ${\{ \Theta^l, \hat{P}(h^l=k), \hat{P}(h^l=k|h^{l'}=k')  \}}_{l=1, k=1,k'=1}^{L,K^l,K^{l'}}$,
$Z$ - number of allowed nodes per layer.
 \textbf{\small{Output:}}{
 \small{An inference graph $G=(N,E)$, where $N$ and $E$ hold clusters (nodes) and their weighted connections (edges) in the graph, respectively

  }} 

 \textbf{\small{Initialization:}} 
 Push $\Omega$ through the network model to get ${\{ Q^l_{t,s}(\Omega)\}}_{l=1}^{L-1}$ (Eq.~\ref{eq:Q_t,s(omega)}) and clusters ${\{C^l_i\}}_{l=1,i=1}^{L,K^l}$.
~{\small
Set $S=m$, $N=C^L_m$, and $E=\emptyset$ 
 
 {\small For $l=L-1,\dots, 1$ }
 

 {\small{{$~~~$ For $t=1,\dots,K^{l}$}}}, {\small{compute $S^{l}(\Omega,S,t)$ (Eq.~\ref{eq:OmageScore})}}

 {\small{{$~~~$ Choose $(z^{l}_1,\dots,z^{l}_Z)$ to be the $Z$ clusters indices with the largest scores $S^{l}(\Omega,S,t)$  }}}

 {\small{$~~~$ Set $S= (z^{l}_1,\dots,z^{l}_Z)$ and  $e^{l}_{i,j} = S^{l}(\Omega,z^{l+1}_i,z^{l}_j)$, $\forall i,j=1,\dots Z$ } }
 
 {\small{$~~~$ Set $N=N\cup{\{C^l_{z^l_i}\}}_{i=1}^Z$ and $E=E\cup{\{e^l_{i,j}\}}_{i=1,j=1}^{Z,Z}$ }}// nodes and edges update

  }}} }
 \end{algorithm}
\subsubsection{Explaining multiple words and images:}
\label{subsec:MWandI}

The optimization problem presented in Eq.~\ref{eq:optML} can be extended to
multiple visual words in multiple images using column position and image independence assumptions.  
Assume a set of validation images $\Omega$ is being analyzed, and a set of words $S \subset \{1,\dots,K^l\}$ from layer $l$ has to be explained by lower layer words for these images. We would like to maximize the likelihood of the set of all column activities $\{h_p^l(I_n): h_p^l \in S, I_n\in \Omega \}$, in which a word from $S$ appears. Assuming column position independence, this likelihood decomposes into terms similar to Eq.~\ref{eq:SingleWord}:


\begin{gather}
\label{eq:optimizationProb_multipleWords}
\scalebox{0.95}{$\begin{align}
   & \log P\Big( \big\{h_p^l(I_n): h_p^l(I_n) \in S , I_n \in \Omega \big\} \big| \big\{ h_q^{l'}(I_n) : I_n \in \Omega  \big\} \Big)  =  \\ \nonumber
  &\sum_{n=1}^{N} \sum_{s \in S} \sum_{\{p : h_p^l(I_n) =s\}} \log P \big( h_p^l(I_n) \big| h_q^{l'}(I_n) , q \in R(p)  \big).\end{align}$}
\end{gather}
Repeating the derivation also given in Eqs.~\ref{eq:SingleWord}, \ref{eq:A+C*log(lr)}, and \ref{eq:optML} for this expression, we get a similar optimization problem,
\begin{equation}
\label{eq:optML2}
\scalebox{0.95}{$
\max_{\underset{|T|=Z}{T}}\left\{ \sum_{t \in T} \sum_{s \in S} Q_{t,s}^{l'}(\Omega)  \log \frac{P\big(h_q^{l'}=t | h_p^l=s, q \in R(p)\big)}{P(h_q^{l'}=t)}\right\} ,$}
\end{equation}
where $Q_{t,s}^{l'}(\Omega)$ is the aggregation of $Q_t^{l'}(I,p)$ over multiple positions and images
\begin{equation}\label{eq:Q_t,s(omega)}
\scalebox{0.95}{$
Q_{t,s}^{l'}(\Omega)  = \sum_{n=1}^N  \sum_{\{p : h_p^l(I_n) =s\}}  Q_t^{l'}(I_n,p).$}
\end{equation}
That is, $Q_{t,s}^{l'}(\Omega)$ is the number of occurrences of word $s$ with word $t$ in its receptive field in all the images in $\Omega$. The solution is given by choosing the $Z$ words in layer $l'$ for which the score
\begin{equation}
\label{eq:OmageScore}
\scalebox{0.95}{$
 S^{l'}(\Omega,S,t) = \sum_{s \in S} Q_{t,s}^{l'}(\Omega)  \log \frac{P\big(h_q^{l'}=t | h_p^l=s, q \in R(p)\big)}{P(h_q^{l'}=t)}$}
\end{equation}
is maximized. 
The inference graph is generated by going over the layers backwards, from the top layer, for which the decision has to be explained, and downwards towards the input layer, selecting the explaining nodes using the score of Eq.~\ref{eq:OmageScore}. See Algorithm~\ref{alg:GraphBuild} for details.



\section{Results}\label{sec:results}

The HMM for MLP formalism was tested by training a fully connected network on the 
CIFAR10~\cite{krizhevsky2009learning} dataset, containing $10$ classes. The network included six layers with the first five containing $1,000$ neurons each.
Based on a preliminary evaluation, the number of visual words $K^l$ was set at $40$ for all layers.

CNN models included ResNet20~\cite{he2016deep} trained on CIFAR10, 
and VGG-16~\cite{simonyan2014very} and ResNet50~\cite{he2016deep} trained on the ILSVRC 2012 dataset~\cite{ILSVRC15}. For ResNet20, the output of all add-layers after each skip connection were modeled, as these outputs are expected to contain aggregated information. For ResNet50 there are $16$ add layers and the output of add-layers 3, 7, 13, and 16 were modeled. For VGG-16, the first convolutional layers at each block were modeled (four layers in total). The numbers of visual words were set at $100,200,450$, and $1,500$ for Layers $1$-$4$, respectively, 
according to the GPU memory limitation.

In all experiments and modeled layers, the GMM's mean parameters were initialized using $K^l$ randomly selected examples. The variance parameters were initialized as the variances computed from $1,000$ random examples. Prior probabilities were uniformly initialized to be $\frac{1}{K^l}$.

\begin{figure}[t]

\begin{center}
\begin{tabular}{c c |c |c |c| c}
\footnotesize Car & &\footnotesize fc-2 & \footnotesize fc-3 & \footnotesize fc-4 \\

\raisebox{-.5\height}{\includegraphics*[width=0.22in,keepaspectratio=true]{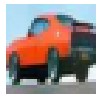}}&  $\rightarrow$ &

\raisebox{-.5\height}{\includegraphics*[width=0.9in,keepaspectratio=true]{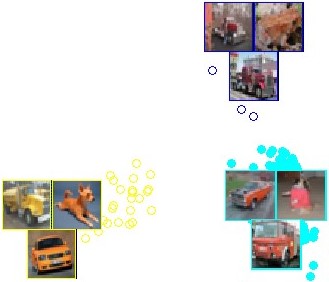}}  & 

\raisebox{-.5\height}{\includegraphics*[width=1.3in,keepaspectratio=true]{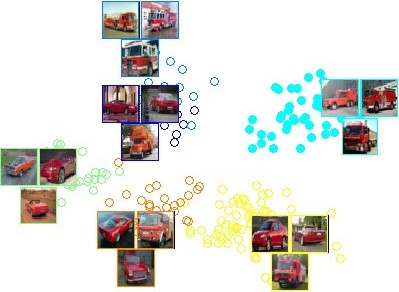}}  &

\raisebox{-.5\height}{\includegraphics*[width=1.0in,keepaspectratio=true]{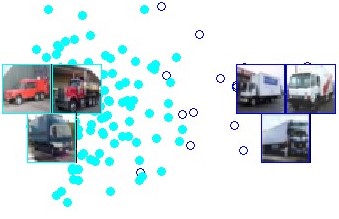}} &

$\rightarrow$ \footnotesize{Truck}
\end{tabular}
\end{center}
\begin{flushleft}
\caption{\footnotesize {\bf MLP inference path}. Three main decision junctions of a misclassified example in a 6-layer network. Subclusters are visualized by the three most representative examples. Points from the subcluster chosen by this example are marked by full circles. 
} \label{fig:Sequential_path}
\end{flushleft}

 \end{figure}
 
\subsection{MLP Inference Path}
\label{sec:MLPres}
In MLP networks, the entire layer activity is assigned to a single cluster. 
To visualize such a cluster, we consider it as a ``decision junction", where a decision regarding the consecutive layer cluster is made.   
For this visualization, the activity vectors in the $C^l_k$ cluster are labeled according to their consecutive layer clusters, forming subclusters for this cluster. 
We use linear discriminant analysis (LDA)~\cite{fukunaga1990introduction} to find a 2D projection of the activity vectors that maximize the separation of the examples with respect to their subcluster labels. Each subcluster is visualized using the three examples with the minimal $l_2$ distance to the subcluster center.

The inference path for an example $I$ is defined to be the maximum a-posteriori (MAP) cluster sequence, i.e., $H=(h^1,...,h^L)$, satisfying 
    \begin{equation}
    \max_{h^1,...,h^L} \log{P(h^1,...,h^{L} | X(I) ) },
    \label{eq:Viterby}
    \end{equation}
where $H$ can be found using the Viterbi algorithm~\cite{forney1973viterbi}.

Such inference paths are useful for error diagnosis. In Fig.~\ref{fig:Sequential_path}, a partial path containing three decision junctions of an erroneous ``car'' example in the CIFAR10 network is presented. 
It can be seen that the example's likely ``decisions" in layer fc-$3$ leads to car and truck clusters in the consecutive layer. At this point, due to its unconventional rear appearance, resembling a truck front, this car example was wrongly associated with a ``truck'' subcluster, an association that remained until the classification layer.

\subsection{Cluster Similarity Across Layers}
\label{sec:ClusterSim}
\begin{figure}[t]
\begin{tabular}{c c}
\begin{tabular}{c c}

\raisebox{3.5\height}{\footnotesize{(a)}} & \includegraphics[width=6.2cm,height=1.7cm]{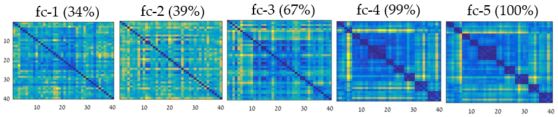}\\
\raisebox{-0.7\height}{\footnotesize (b)} & \raisebox{-.5\height}{\includegraphics[width=6.2cm,height=1.7cm]{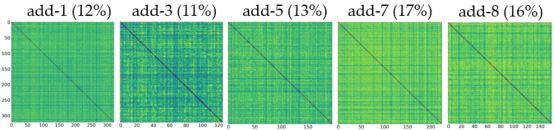}}
\end{tabular}
&
\hspace{2mm}
\raisebox{-.5\height}{\includegraphics[width=4.8cm,height=4cm]{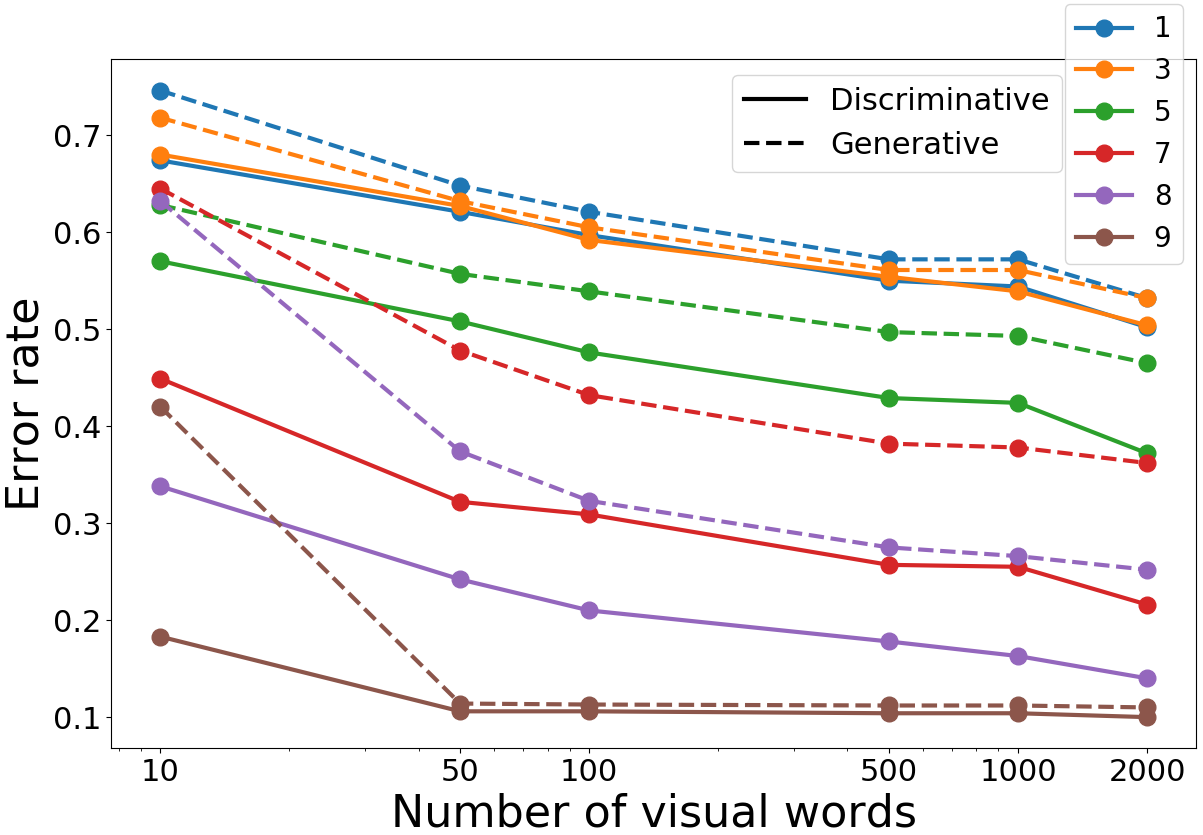}}

\end{tabular}
\begin{flushleft}

\caption{\footnotesize {\bf Left:} Cluster similarity matrices for increasing layer indices in a 6-layer MLP (a) and ResNet20 (b) both trained on CIFAR10. The average percentage of dominant class examples (across clusters) is stated above each matrix.
{\bf Right:} Error rates of a linear classifier trained over word histograms taken from six ResNet20 conv-layers (1, 3, 5, 7, 8, and 9) trained with either a generative or discriminative loss (Section~\ref{subsubsec:CNN}).
}
 \label{fig:distance_matrix}
\end{flushleft} 
 \end{figure}

A plausible assumption about network layer representations is that early layers are input dominated, while representations in late layers are more class-related. This phenomenon can be observed in our framework by considering how distance between clusters changes as a function of layer index.  
In Fig.~\ref{fig:distance_matrix} (left), similarity matrices are presented. Each matrix shows Euclidean distances between centers of the clusters from a single layer. Clusters are ordered by their dominant class index,  
defined as 
the class whose examples are the most frequent in the cluster. 

For an MLP network, presented in Fig.~\ref{fig:distance_matrix} (top left), the progression toward class-related representation is evident from the emerging block structure in Layers $3$-$5$, indicating increasing similarity between clusters representing the same class.
In contrast, as seen in Fig.~\ref{fig:distance_matrix} (bottom left), CNN clusters (representing column activities) stay local and diverse, even at the uppermost layers where their receptive fields cover the entire input space.
This phenomenon is demonstrated by the lack of block structure, as well as the relatively low frequency of the dominant class in a cluster.
This indicates that the final CNN classification is based on several class-oriented words, which are not similar, and  appear simultaneously in different image regions.

\subsection{CNN Inference Graphs}
\label{subsec:cnn_res}

\begin{figure}[!tp]
\centering
\begin{tabular}{c}
\normalsize Class pineapple\\

\raisebox{-.5\height}{\includegraphics[width=0.62\columnwidth]{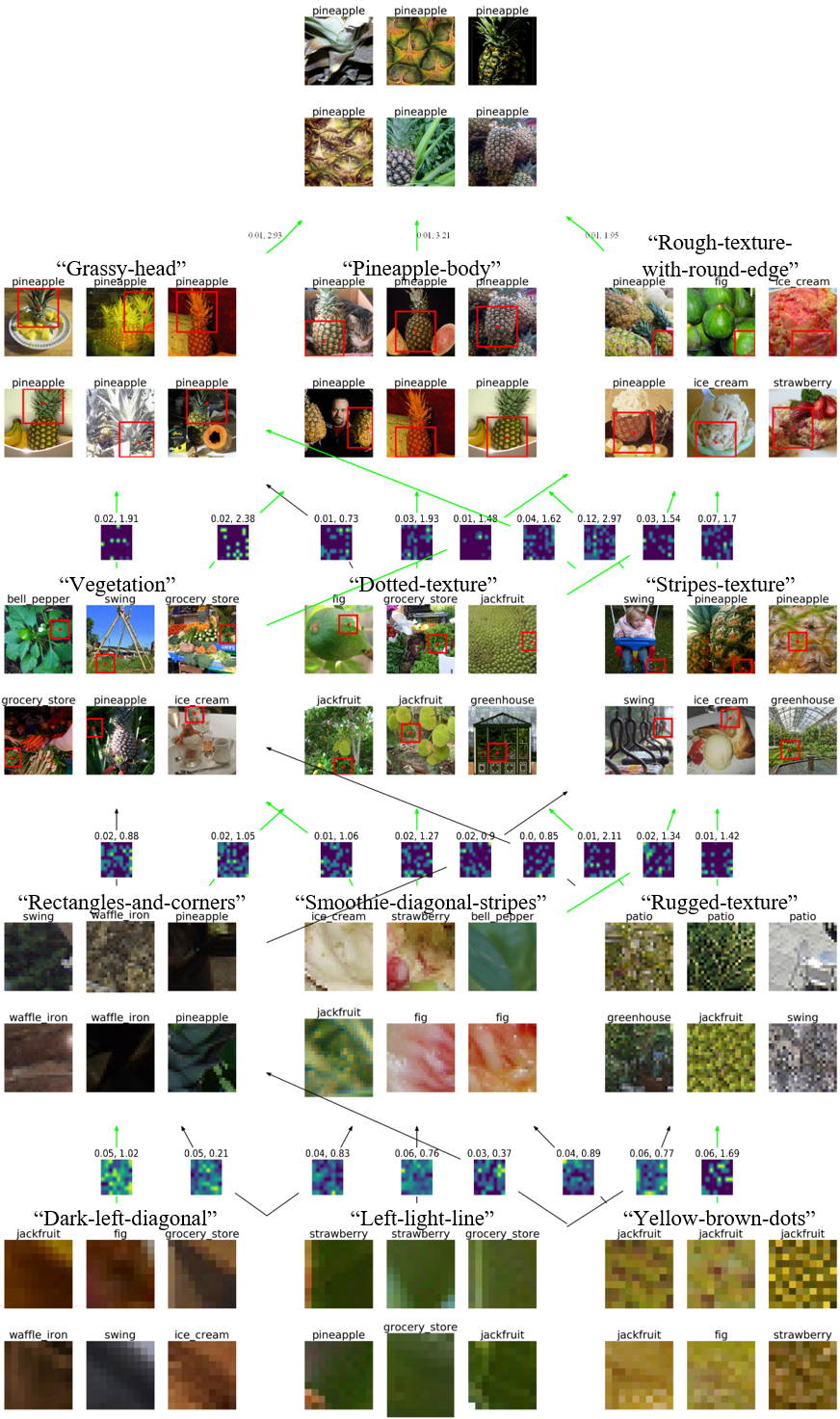}}
\end{tabular}
\caption{ \textbf{Pineapple inference graph}. The graph is generated by training a model on ”pineapple” class and its neighboring classes. The top node is a visual word of the output layer, representing the predicted class ``pineapple''. The lower levels in the graph show the three most influential words in preceding modeled layers. Visual words are manifested by the six representative examples for which $P(h^l=k|x_p^l)$ is the highest. For the two highest layers, examples are presented by showing the example image with a rectangle highlighting the receptive field of the word's location. For lower layers, the receptive field patches themselves are shown. Images are annotated by their true label. Arrows are shown when the log-ratio term is positive, colored green for significant connections in which the term is higher than $1$. They are annotated by the frequency of the lower word in the receptive field (left) and the log ratio (right) (the two components of the score in Eq.~\ref{eq:OmageScore}).
In addition, for each arrow, a heatmap is shown indicating the frequent locations of the lower-level visual words in the receptive field of the higher-level word. Tags above each cluster were added by the authors for figure explanation convenience. The figure is best inspected by zooming in on clusters of interest.
}
\label{fig:pineapple_tree}
 \end{figure}

\begin{figure}[!tp]
\begin{center}
\begin{tabular}{c}
\normalsize Class swing\\

\raisebox{-.5\height}{\includegraphics[width=0.86\columnwidth]{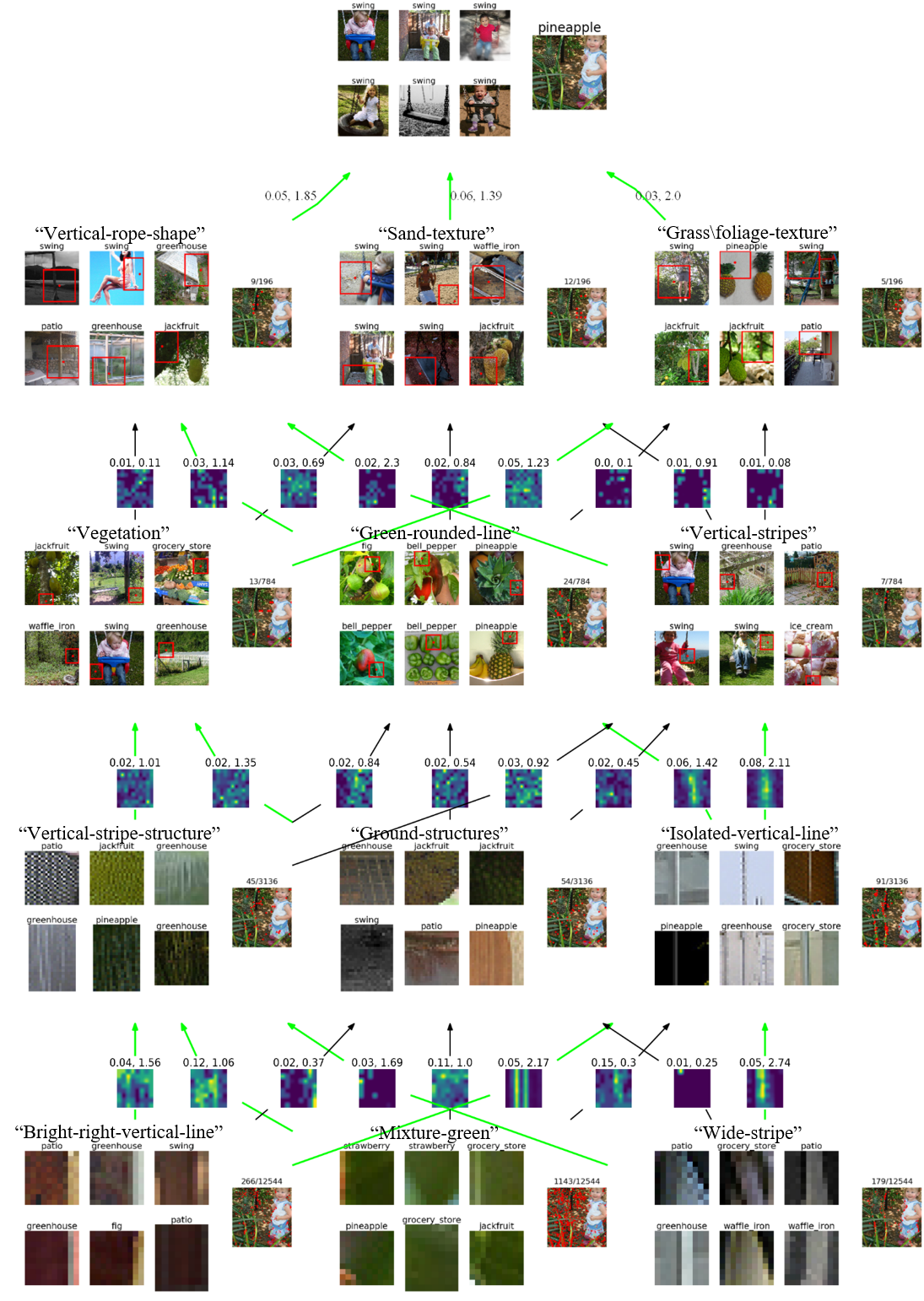}}
\end{tabular}
\end{center}

\caption{ \textbf{An image inference graph of an erroneous image}. An image inference graph for a pineapple image wrongly classified to the class "swing''. The model is generated using "pineapple" and its neighboring classes (same as in Fig~\ref{fig:pineapple_tree}), 
where the neighbor class "swing'' is included.
The graph is generated by applying the node selection algorithm (Section~\ref{subsec:node_selelction}) to a set $\Omega$ containing this single erroneous image.
The analyzed image is shown on the right side of each cluster node, with red dots marking spatial locations assigned to the cluster. 
The fraction of spatial examples belonging to the cluster (in this image) appears in the title. 
}
\label{subfig:bad_image_tree_pineapple}

\end{figure}
 
 \begin{figure}[t]
\centering
\begin{tabular}{c c}
\raisebox{-.4\height}{\includegraphics[width=0.63\columnwidth]{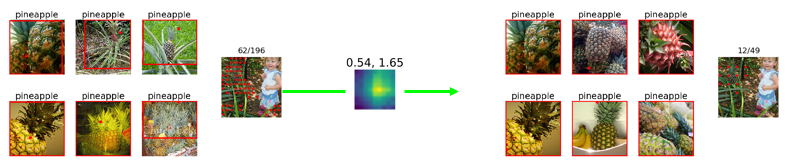}}&
$\longrightarrow$ \footnotesize{Class pineapple}
 
\end{tabular}
\begin{flushleft}

\caption{\footnotesize {\bf Subgraph of the same image as in Fig.~\ref{subfig:bad_image_tree_pineapple} classified by ResNet50.} A visual word in layer add-13 (left), leading to a visual word in layer add-16 (middle), and to correct classification of the pineapple (right).  
} \label{fig:resnet_inference_trees}
\end{flushleft} 
 \end{figure}

{\bf Loss and dictionary size.}
Fig.~\ref{fig:distance_matrix} (right) shows the errors obtained for linear classification using dictionary histograms (Eq.~\ref{eq:discriminative_loss}), as a function of dictionary size. Graphs are shown for dictionaries obtained using losses $\mathcal{L}_{G}$ (\ref{eq:generative_loss}) and $\mathcal{L}_{D}$ (\ref{eq:discriminative_loss}), for several intermediate convolutional layers of the CIFAR10 network. For all layers, except the final one (indexed 9), larger dictionaries provide better classification. However, for the final layer, dictionaries larger than $50$ clusters do not increase accuracy, which approaches the original network error of $0.088$.
As expected, the discriminative loss $\mathcal{L}_{D}$ leads to smaller errors than the generatively-optimized loss, as the former directly minimizes the classification error. Therefore, all models presented below were trained with the $\mathcal{L}_{D}$ loss.

\textbf{Class inference graph.}
An example of a class inference graph for the class ``pineapple'' in VGG-16 is presented in Fig.~\ref{fig:pineapple_tree}. 
The graph shows that the most influential words in the top convolution layer can be roughly characterized as ``Grassy-head'', ``Pineapple-body'', and ``Rough-textured-with(lower)-round-edge''. The origin of these words can be traced back to lower layers. For example, ``Grassy-head'' is composed of words capturing mostly ``Vegetation'' in the layer below, which are in turn generated from words describing green texture and multiple diagonal lines (in Layer $2$). Similarly, the origin of ``Pineapple-body'' can be traced back to yellow and brown texture words in lower layers.


\textbf{Image inference graphs}.
Fig.~\ref{subfig:bad_image_tree_pineapple} shows an image inference graph for a pineapple image wrongly classified to the ``swing'' class. Using the inference graph, we can analyze the dominant (representative) visual words that have led to this erroneous classification:
\textbf{(1)} Top layer (Layer $4$): The visual words voting for the swing class focus on strong ``Vertical-rope-shape", ``Sand-texture", and ``Grass/foliage-texture". 
\textbf{(2)} Layers $3$ and $2$: 
The ``Vertical-rope-shape" (of Layer $4$) originates from a similar visual word, "Vertical-stripes", of Layer $3$, and this in turn depends strongly on the ''Isolated-vertical-line'' word in Layer $2$. The foliage word (in Layer $4$) mainly originates from the ``Vegetation" word in Layer $3$, which in turn heavily depends on the two brown/green ``vertical-stripe-structure'' and ``Ground-structure" words in Layer $2$.
\textbf{(3)} Layer $1$: the main explanatory words in Layer $1$ are green and bright vertical edges and lines, which are combined to construct the ``Isolated-vertical-line'' and ``Vertical-stripe-structure'' words in Layer $2$.

In Fig.~\ref{fig:resnet_inference_trees}, we show a partial inference graph of the same ``pineapple" image wrongly classified to class ``swing" by VGG-16 (Fig.~\ref{subfig:bad_image_tree_pineapple}), which is successfully classified by ResNet50. As can be seen in the top layer of the graph (add-16, middle), Resnet50 successfully detects the pineapple location in the image, where both visual words presented contain strong ``pineapple" features. 
Additional examples of class and image inference graphs are given in the Supplementary Material.
 

\section{Conclusions}
We introduced a new approach for interpreting hidden layers activity of deep neural networks by learning dictionaries of activity clusters and transition probabilities between clusters of consecutive modeled layers. We formalized a maximum-likelihood criterion for mining clusters relevant for network prediction, which enable building explanatory inference graphs of manageable size. Inference graphs can be constructed for an entire class, to understand the general network reasoning for this class, or for specific images, for which error analysis may specifically be sought. The tools developed can be used to verify the soundness of the network reasoning and to understand its hidden inference mechanisms, or conversely to reveal network weaknesses. 



\noindent \textbf{Acknowledgments:}
This work was supported by the Israeli Ministry of Science and Technology and Israel Innovation Authority through the Phenomics consortium.

\bibliographystyle{splncs}
\bibliography{egbib}
\end{document}